\documentclass{midl} 

\usepackage{mwe} 
\usepackage{float}
\jmlrvolume{-- Under Review}
\jmlryear{2026}
\jmlrworkshop{Full Paper -- MIDL 2026 submission}
\editors{Under Review for MIDL 2026}

\title[Connetcome Hybrid VAE]{Unsupervised learning of acquisition variability in structural connectomes via hybrid latent space modeling}

\midlauthor{
\Name{Gaurav Rudravaram\nametag{$^{1}$}}  \Email{gaurav.rudravaram@vanderbilt.edu}\\
\addr $^{1}$ Department of Electrical and Computer Engineering, Vanderbilt University, Nashville, TN, USA \\
\Name{Lianrui Zuo\nametag{$^{1}$}} \Email{lianrui.zuo@vanderbilt.edu}\\
\Name{Michael E. Kim\nametag{$^{2}$}} \Email{michael.kim@vanderbilt.edu}\\
\addr $^{2}$ Department of Computer Science, Vanderbilt University, Nashville, TN, USA \\
\Name{Karthik Ramadass\nametag{$^{1,2}$}} \Email{karthik.ramadass@vanderbilt.edu}\\
\Name{Elyssa McMaster\nametag{$^{1}$}}
\Email{elyssa.mcmaster@vanderbilt.edu} \\
\Name{Jongyeon Yoon\nametag{$^{2}$}}
\Email{jongyeon.yoon@vanderbilt.ed} \\
\Name{Aravind R. Krishnan\nametag{$^{1}$}}
\Email{aravind.r.krishnan@vanderbilt.edu} \\
\Name{Adam M. Saunders\nametag{$^{1}$}}
\Email{adam.m.saunders@vanderbilt.edu} \\
\Name{Chenyu Gao\nametag{$^{1}$}}
\Email{chenyu.gao@vanderbilt.edu} \\
\Name{Nancy R. Newlin\nametag{$^{3}$}}
\Email{newlinn@mskcc.org} \\
\addr $^{3}$ Memorial Sloan Kettering Cancer Center, New York, NY, USA
 \\
\Name{Praitayini Kanakaraj\nametag{$^{2}$}}
\Email{praitayini.kanakaraj@vanderbilt.ed} \\
\Name{Lori L. Beason Held\nametag{$^{4}$}}
\Email{heldlo@grc.nia.nih.gov} \\
\addr $^{4}$ Laboratory of Behavioral Neuroscience, National Institute on Aging, National Institutes of Health, Baltimore, MD, USA \\
\Name{Murat Bilgel\nametag{$^{4}$}}
\Email{murat.bilgel@nih.gov} \\
\Name{Laura A. Barquero\nametag{$^{5}$}}
\Email{laura.barquero@Vanderbilt.Edu} \\
\addr $^{5}$ Peabody College of Education and Human Development, Nashville, Tennessee, USA
 \\
\Name{Micah D’Archangel\nametag{$^{5}$}}
\Email{micah.a.darchangel@vanderbilt.edu} \\
\Name{Tin Q. Nguyen\nametag{$^{5}$}}
\Email{tin.nguyen@vanderbilt.edu} \\
\Name{Laurie B. Cutting\nametag{$^{5}$}}
\Email{laurie.cutting@Vanderbilt.Edu} \\
\Name{Derek Archer\nametag{$^{6}$}}
\Email{derek.archer@vumc.org} \\
\addr $^{6}$Vanderbilt University Medical Center,Vanderbilt Memory $\&$ Alzheimer’s Center,Nashville,TN, USA \\
\Name{Timothy J. Hohman\nametag{$^{6}$}}
\Email{timothy.j.hohman@vumc.org} \\
\Name{Daniel C. Moyer\nametag{$^{2}$}}
\Email{daniel.moyer@Vanderbilt.Edu} \\
\Name{Bennett A. Landman\nametag{$^{1,2}$}}
\Email{bennett.landman@vanderbilt.edu} \\
}

\begin{document}

\maketitle

\begin{abstract}
Acquisition differences across sites, scanners, and protocols in dMRI introduce variability in structural connectome analysis. 
This motivates the need for deep learning models that can represent downstream, high-dimensional structural connectomes in a low-dimensional space while explicitly separating acquisition-related effects from underlying biological variation.
Conventional statistical and deep learning approaches for dimensionality reduction typically model all sources of variance as continuous, making it difficult to separate discrete effects, e.g., acquisition- or site-related, from continuous biological variation. 
As a result, acquisition-related effects often become absorbed into a continuous latent space. 
Recent advances in deep learning have explored hybrid latent space modeling, where discrete and continuous components jointly represent structured variability. 
However, existing hybrid approaches generally rely on manual capacity tuning to ensure that the discrete component captures desired variability (e.g., acquisition).
Here, we introduce a principled unsupervised framework that removes the need for such manual capacity tuning by \textit{architecturally annealing} the encoder outputs before decoding, allowing the model to adaptively balance the contributions of discrete and continuous latent variables during training. 
To investigate this joint latent space modeling, we curated a large dataset ($N=7,416$; 60\% female) of structural connectomes derived from dMRI scans of participants.
Our dataset spans an age range of 2 to 102 years and encompasses 13 different studies with 25 unique acquisition parameter combinations. 
Among these, 5,900 are cognitively unimpaired/neurotypical, 877 are diagnosed with mild cognitive impairment (MCI), and 639 are diagnosed with Alzheimer’s disease (AD). 
We compare our approach with a standard VAE, PCA followed by k-means clustering, and hybrid models that impose annealing only through the loss function, showing that the architectural annealing results in stronger site learning (ARI=$0.53$, $p < 0.05$) as compared to the other methods. These results demonstrate that the proposed hybrid continuous–discrete latent space provides a useful unsupervised mechanism for capturing acquisition-related variability in diffusion MRI; by jointly modeling smooth and categorical structure, the Joint-VAE recovers meaningful clusters aligned with scanner and protocol differences.
\end{abstract}

\begin{keywords}
Unsupervised representation learning, structural connectomes, harmonization
\end{keywords}

\section{Introduction}
White-matter connectivity and microstructural integrity are central to understanding neurodegenerative diseases~\cite{kamagata2021diffusion}, cognitive decline~\cite{vogt2020cortical}, and aging~\cite{kantarci2014white}.
Diffusion MRI (dMRI) is well suited for these investigations because water molecule diffusion is constrained by axonal membranes and myelin, providing indirect signatures of microstructural organization~\cite{jones2010diffusion}.
By modeling diffusion‐weighted signals, dMRI enables the estimation of quantitative metrics such as fractional anisotropy (FA), mean diffusivity (MD), and more advanced compartment-based measures that reflect axonal integrity and tissue composition~\cite{mori2013introduction}.
These signals can also be used to perform tractography and construct structural connectomes that characterize large-scale network organization for downstream statistical or machine-learning analyses~\cite{shamir2025tutorial}.

Despite its importance, dMRI is heavily impacted by heterogeneity in acquisition, which complicates the ability to draw generalizable conclusions across studies or cohorts. 
Diffusion-derived metrics have been shown to vary significantly with echo time~(TE), repetition time~(TR), magnetic field strength, and diffusion weighting~\cite{ hui2010b, yao2023both}. 
Higher b-values reduce signal-to-noise ratio, and both~\citet{hui2010b} and~\citet{yao2023both} demonstrated that FA, MD, and related quantities differ substantially when computed from data acquired at different b-values, reflecting a combination of noise-floor effects and underlying tissue-compartment differences. 
Multi-shell acquisitions introduce additional complexity: while they enable more expressive modeling of crossing fibers and multi-compartment microstructure---supporting models such as NODDI~\cite{zhang2012noddi, jelescu2017design}---they also produce connectomes that differ meaningfully from those derived from single-shell data, with \citet{yao2023both} showing improved sensitivity for predicting Parkinson’s disease when using multi-shell–derived connectomes. 
Although such acquisition differences are often intentionally leveraged to optimize sensitivity to particular microstructural properties or brain regions \cite{caiazzo2016q}, they become problematic when integrating data across scanners or protocols: even subtle variations alter the diffusion signal and propagate to microstructural estimates, tractography, and ultimately the connectomes themselves. 
Consequently, connectomes generated from the same individual under different acquisition protocols can exhibit substantial discrepancies \cite{prvckovska2016reproducibility, villalon2016reliability}, making it difficult to distinguish biological effects from acquisition-driven variability in multi-site analyses.

This variability has motivated the development of harmonization techniques aimed at reducing acquisition-related differences before group analyses or machine-learning tasks. 
Statistical approaches such as ComBat\cite{fortin2017harmonization} have been applied to remove protocol effects from scalar brain connectivity metrics, and recent deep-learning methods have sought to learn latent representations of structural connectomes that are invariant to acquisition domains.
Autoencoders \citep{zheng2025connectomeae}, graph neural networks \citep{noman2024graph}, and conditional variational autoencoders~(CVAEs) \citep{newlin2025harmonizing, zuo2021unsupervised, zuo2023haca3} have all been used to enforce site invariance by conditioning on site labels or acquisition-related variables. 
Adversarial graph neural network frameworks have further incorporated site classifiers and site-conditioned decoders to suppress domain information~\citep{patel2025structural}. However, these approaches all depend on predefined site or scanner labels, a significant limitation given that acquisition variability may exist within a single dataset or metadata may be incomplete or inaccurate \citep{zuo2022disentangling}. 
This motivates the need for unsupervised models capable of disentangling acquisition-related variation directly from the data itself.

Hybrid latent-space models that combine continuous and discrete representations offer a promising path toward such unsupervised disentanglement. Following the insight from~\citet{dupont2018learning} that discrete latent variables can capture structured, categorical variation, we treat site effects (arising from combinations of acquisition parameters and protocol differences) as a discrete component superimposed on continuous biological variation. 
This joint representation allows the model to infer site structure without any site labels~\citep{rudravaram2025characterizing}. 
Yet the original hybrid formulation relies on a heuristic hinge-margin mechanism to balance capacity between continuous and discrete latent spaces, providing only indirect control over information allocation. 
Our prior work \citep{rudravaram2025characterizing} addressed this by introducing a loss-based capacity regulation strategy that improved stability and interpretability.
In this paper, we extend the framework to a substantially larger and more heterogeneous multi-site dataset and introduce a staged, model-based annealing procedure that adjusts capacity within the architecture itself. 
This new strategy increases discrete-space utilization and enables recovery of more meaningful acquisition-related clusters in a fully unsupervised manner.

\section{Methods}
\subsection{Data and preprocessing}
To evaluate hybrid continuous-discrete latent space models for structural connectomes, we assembled a large multi-cohort dMRI dataset spanning 13 major neuroimaging studies. 
For datasets with longitudinal imaging, we selected a single scan per participant to avoid bias from repeated measures. 
The final dataset comprises 955 connectomes from BLSA \cite{ferrucci2008baltimore}, 57 from ADNI, 87 from the Calgary Preschool dataset \cite{reynolds2020calgary}, 373 from a pediatric VUMC dataset \cite{ds004146:1.0.0}, 111 from the Multisensory Lexical Processing cohort \cite{ds001894:1.4.2}, 80 from MASiVar \cite{ds003416:2.0.2}, 236 from a longitudinal language-development study \cite{ds003604:1.0.7}, 398 from QTAB\cite{ds004146:1.0.3}, 2,486 from HABSHD \cite{petersen2025health}, 339 from WRAP \cite{johnson2018wisconsin}, 1,006 from NACC, and 610 from a combined ROS/MAP/MARS cohort \cite{bennett2005rush,a2012overview,l2012minority}. 
Across datasets, we defined a site as a unique scanner–protocol pairing, yielding 25 distinct sites. 
The aggregated sample includes 4,490 females and 2,926 males (ages $2 \sim 102$), consisting of neurotypical participants ($n = 5,900$) as well as individuals diagnosed with mild cognitive impairment (MCI; $n = 877$) and Alzheimer’s disease (AD; $n = 639$).

All imaging data were preprocessed following \citet{kim2025scalable}. 
Diffusion MRI scans were corrected and standardized using the PreQual pipeline \cite{cai2021prequal}. 
Structural MRI volumes were segmented into 121 BrainColor regions using the SLANT brain segmentaton \cite{huo20193d}. 
Fiber orientation distributions were estimated and whole-brain probabilistic tractography was performed in MRtrix3 \cite{tournier2012mrtrix}, seeding streamlines at the white matter-gray matter interface and generating 10 million streamlines per subject. Structural connectomes were constructed by counting streamline terminations between all SLANT-defined ROI pairs. 
Visual quality assurance was performed at each processing stage, following best practices in \citet{kim2025scalable}.

\subsection{JointVAE}
\textbf{Traditional variational autoencoders (VAEs)} consist of an encoder $q_\phi(\cdot)$, parameterized by $\phi$, that maps the input data $x$ into a low-dimensional latent representation $z$, and a decoder~$p_\theta(\cdot)$, parameterized by $\theta$, that reconstructs the input as $\hat{x}$. To prevent the latent space from becoming arbitrarily dispersed or fragmented, the approximate posterior $q_{\phi}(z | x)$ is encouraged to match a prior distribution, typically a standard normal distribution $p(z) = \mathcal{N}(0, I)$. 
Under this formulation, the traditional VAE optimization objective becomes:
\begin{equation}
\mathcal{L}(\theta, \phi)
= \mathbb{E}_{q_{\phi}(z | x)}\!\left[\log p_{\theta}(x | z)\right]
- D_{\mathrm{KL}}\left(q_{\phi}(z | x)\,\|\, p(z)\right).
\end{equation}
\textbf{The Joint-VAE}~\cite{dupont2018learning} extends the traditional VAE framework by augmenting the latent space with an additional discrete component. 
Let $z_c$ denote the continuous latent variable and $z_d$ denote the discrete latent variable.
The encoder now estimates a joint posterior distribution, $q_{\phi}(z_c, z_d | x)$, while the decoder reconstructs the input from both latent components via $p_{\theta}(x | z_c, z_d)$. 
With this formulation, the Joint-VAE objective becomes:
\begin{equation}
\mathcal{L}(\theta, \phi)
= \mathbb{E}_{q_{\phi}(z_c, z_d | x)}\!\left[\log p_{\theta}(x | z_c, z_d)\right]
- D_{\mathrm{KL}}\left(q_{\phi}(z_c, z_d | x)\,\|\, p(z_c, z_d)\right).
\end{equation}
Because $q_{\phi}(z_c \mid x)$ and $q_{\phi}(z_d \mid x)$ are produced as separate distributions from the encoder, the posterior can be factorized as $q_{\phi}(z_c, z_d \mid x) = q_{\phi}(z_c \mid x)\, q_{\phi}(z_d \mid x)$, and with the standard factorized prior $p(z_c, z_d) = p(z_c)\, p(z_d)$ the KL divergence decomposes into two terms.
\begin{equation}
D_{\mathrm{KL}}\!\left(q_{\phi}(z_c, z_d | x)\,\|\, p(z_c, z_d)\right)
= D_{\mathrm{KL}}\!\left(q_{\phi}(z_c | x)\,\|\, p(z_c)\right)
+ D_{\mathrm{KL}}\!\left(q_{\phi}(z_d | x)\,\|\, p(z_d)\right).
\end{equation}
Thus, the final objective for the Joint-VAE training becomes:
\begin{equation}
\mathcal{L}(\theta, \phi)
= \mathbb{E}_{q_{\phi}(z_c, z_d | x)}\!\left[\log p_{\theta}(x | z_c, z_d)\right]
- \beta_c\, D_{\mathrm{KL}}\!\left(q_{\phi}(z_c | x)\,\|\, p(z_c)\right)
- \beta_d\, D_{\mathrm{KL}}\!\left(q_{\phi}(z_d | x)\,\|\, p(z_d)\right),
\end{equation}
where $\beta_c$ and $\beta_d$ allow separate weighting of the continuous and discrete KL terms.
However, because the continuous latent space has, in principle, unbounded capacity compared to the discrete space, optimizing this loss directly often leads the model to place almost all information in the continuous latent $z_c$ and ignore the discrete latent $z_d$.

To address this imbalance, the original Joint-VAE \cite{dupont2018learning} proposes gradually increasing the ``capacity'' of each latent channel throughout training so that the continuous and discrete parts can contribute at different rates. 
This is implemented by introducing hinge parameters $C_c$ and $C_d$, which specify target capacities (upper bounds) for the KL contributions of the continuous and discrete latent spaces, respectively, and are treated as user-defined hyperparameters. 
The resulting objective is
\begin{equation}
\begin{aligned}
\mathcal{L}(\theta, \phi)
&= \mathbb{E}_{q_{\phi}(z_c, z_d | x)} \!\left[ \log p_{\theta}(x | z_c, z_d) \right]  \\
&\quad - \beta \Bigl| D_{\mathrm{KL}}\!\bigl(q_{\phi}(z_c \mid x)\,\|\, p(z_c)\bigr) - C_c \Bigr|
      - \beta \Bigl| D_{\mathrm{KL}}\!\bigl(q_{\phi}(z_d | x)\,\|\, p(z_d)\bigr) - C_d \Bigr|.
\end{aligned}
\label{eq:original_joint_vae}
\end{equation}
In this hinge-based formulation, the loss now depends on three hyperparameters: the continuous and discrete capacities $C_c$ and $C_d$, and the weighting factor $\beta$. 
Unlike in the traditional $\beta$-VAE setting, where increasing $\beta$ can encourage disentanglement in a principled evidence lower bound(ELBO) framework, here $\beta$ primarily controls how strictly the model is forced to match the manually specified capacities. 

\subsection{Improved Joint-VAE with Loss and Architecture Annealing}
\textbf{Loss Annealing.} To address this this, in our prior work~\cite{rudravaram2025characterizing}, we proposed a principled staged capacity-control mechanism designed specifically for hybrid continuous-discrete latent spaces. 
Rather than adopting the KL-capacity formulation of \citet{burgess2018understanding}, we directly regulated the expressive power of the continuous latent by annealing the encoder's posterior parameters. 
Let $(\mu, \log\sigma^2)$ denote the encoder outputs for the continuous latent $z_c$. We defined annealed parameters $\mu' = \lambda\, \mu$ and $\log\sigma'^2 = \lambda\, \log\sigma^2$,
where the annealing coefficient $\lambda$ increases linearly from $0$ to $1$ over a fixed number of iterations. Substituting $(\mu', \log\sigma'^2)$ into the KL divergence yields an annealed continuous KL term,
\begin{equation}
\widetilde{D}_{\mathrm{KL}}\!\left(q_{\phi}(z_c \mid x)\,\|\, p(z_c)\right)
= \tfrac{1}{2}\bigl(\sigma^{2\lambda} + \lambda^{2}\mu^{2} - 1 - \lambda \log\sigma^{2}\bigr),
\end{equation}
and the corresponding objective becomes
\begin{equation}
\mathcal{L}(\theta,\phi)
= \mathbb{E}_{q_{\phi}(z_c,z_d | x)}\!\left[\log p_{\theta}(x | z_c,z_d)\right]
- \beta\,\widetilde{D}_{\mathrm{KL}}\!\left(q_{\phi}(z_c | x)\,\|\, p(z_c)\right)
- \beta\,D_{\mathrm{KL}}\!\left(q_{\phi}(z_d | x)\,\|\, p(z_d)\right).
\label{eq:loss_anneal_vae}
\end{equation}

This annealing schedule reduces the influence of the continuous latent at early iterations, encouraging the model to encode structure in the discrete latent $z_d$ before gradually incorporating the continuous latent as $\lambda$ increases. 
As $\lambda$ approaches $1$, the continuous channel becomes fully expressive, enabling balanced joint use of both latent spaces. 
While this loss-based annealing improved stability and reduced dependence on manual capacity tuning, it has a key limitation: the annealing is applied only to the KL term, not to the encoder itself. 
The raw encoder outputs $(\mu, \log\sigma^2)$ remain unconstrained and can drift toward extreme values during early epochs when the KL penalty is weak, leading to unstable dynamics or deviations from the intended capacity schedule. 
Freezing or ignoring the continuous KL term is also insufficient, as these parameters may drift into degenerate regions of parameter space while unregularized.

\textbf{Architectural Annealing.} In this paper, we introduce a novel \emph{architectural annealing} strategy that enforces the capacity schedule at the encoder level. 
Instead of annealing only the KL divergence, we anneal the encoder outputs themselves and use $\mu_\lambda \Leftarrow \lambda\, \mu$ and $\log\sigma^2_\lambda \Leftarrow \lambda\, \log\sigma^{2}$ for both sampling and KL computation. 
Because the decoder receives latent samples drawn from this annealed posterior, it is conditioned throughout training on representations whose capacity is directly controlled. 
This modification ensures that the influence of the continuous latent is minimal at the start of training and then increases smoothly and predictably as $\lambda$ grows. 
By constraining the encoder outputs—and consequently the decoder inputs, rather than modifying the loss alone, the model is forced to rely on the discrete latent early in training and transitions gradually to full joint optimization, avoiding the instability and failure modes associated with loss-based annealing.
\begin{figure}[H]
\floatconts
  {fig:methods_figure}
  {\caption{We encode each flattened connectome $X$ into a mean, log-variance, and discrete class probabilities. An annealing factor (ramping from $0$ to $1$) scales the mean and log-variance during training to suppress the continuous pathway early on and encourage reliance on the discrete space before transitioning to full joint optimization. The continuous latent variable $z_c$ is sampled via the VAE reparameterization trick, and the discrete variable $z_d$ is sampled using the Gumbel--Softmax followed by an $\arg\max$. The concatenated latent vector $[z_c, z_d]$ is then decoded to reconstruct $\hat{X}$.
}}
  {\includegraphics[width=0.8\linewidth]{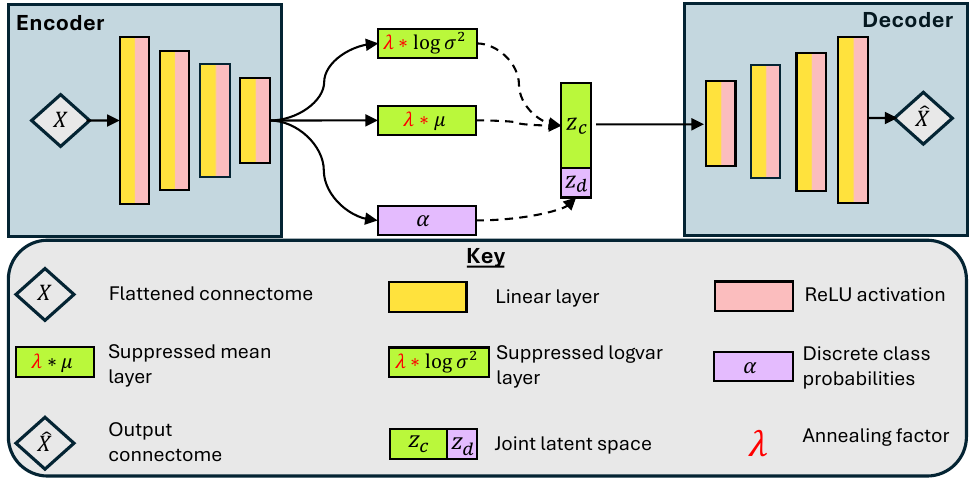}}
\end{figure}
\subsection{Framework of Architectural Annealing Joint-VAE}
Following this annealing mechanism, our Architectural Annealing Joint-VAE framework~(Figure~\ref{fig:methods_figure}) models structural connectomes using both continuous and discrete latent variables. The encoder receives the flattened upper-triangular portion of each connectome and processes it through four fully connected layers with ReLU activations.
From the final encoder layer, we obtain three outputs: the mean vector $\mu$, the log-variance vector $\log\sigma^{2}$, and the logits $\alpha$ that parameterize the discrete latent. 
The continuous latent $z_c$ is sampled using the standard VAE reparameterization trick applied to the annealed parameters $(\mu_\lambda, \log\sigma^2_\lambda)$, while the discrete latent $z_d$ represents 25 categorical classes and is sampled via the Gumbel-Softmax reparameterization trick. 
The decoder mirrors the encoder's structure and reconstructs the predicted connectome $\hat{X}$ from the concatenated latent vector $(z_c, z_d)$, allowing the model to leverage both latent pathways throughout training. The model is trained for 2500 epochs with a batch size of 512 and each epoch has 14 iterations. The $\lambda$ value was increased from 0 to 1 over $5,000$ iterations and for the rest of the training regime, it remained at 1.

\subsection{Experiments}
In our first experiment, we evaluate whether the manual capacity-tuning strategy from the original Joint-VAE (Eq.~\ref{eq:original_joint_vae}) provides stable control over how information is allocated between the continuous and discrete latent spaces. We fix the penalty weight to $\beta = 100$ to strongly enforce adherence to the prescribed capacities. For the discrete latent space, the KL term is theoretically upper-bounded by $\log K$ under a uniform prior, and \citet{dupont2018learning} recommend setting the target capacity equal to this value; accordingly, we set $C_d = \log K$. We then vary only the continuous capacity, training three models with $C_c \in {50, 500, 2000}$ to assess how sensitive site-structure recovery is to this manual capacity specification.

Next, we compare our proposed model-based annealing strategy against the alternative approach that applies annealing solely in the loss function~(Eq.~\ref{eq:loss_anneal_vae}). 
We include two additional baselines: a linear principle component analysis~(PCA) and $k$-means pipeline and a nonlinear baseline where a standard VAE’s continuous latent space is clustered using $k$-means. For all hybrid models, discrete assignments are obtained from the Gumbel–Softmax output.
To quantify how well each method recovers the underlying acquisition sites, we compute the Adjusted Rand Index (ARI) between inferred clusters and ground-truth site labels. ARI measures similarity between two labelings while correcting for chance agreement, ranging from $–1$ (worse than random) to $1$ (perfect agreement), with 0 indicating chance-level correspondence. This makes ARI well suited for evaluating latent site-structure discovery. To assess variability, we bootstrap the dataset with $1,000$ resamples and report the resulting ARI distributions for all models. We perform a t-test on the bootstrapped ARI values at 25 classes, which is the true number of sites to demonstrate that the observed improvements are statistically significant.
\section{Results}
\subsection{Manual Capacity tuning on Joint-VAE}
When $C_c = 50$, the discrete codes recover a reasonable site structure, with distinct clusters that align well with the true site labels. However, as $C_c$ is increased to $500$ and then $2,000$, the model increasingly relies on the continuous latent $z_c$, and the discrete assignments lose granularity, collapsing previously distinct site clusters~(Figure~\ref{fig:problem_figure}) . This behavior highlights that the Joint-VAE objective is highly sensitive to manual capacity tuning and that small changes in $C_c$ can qualitatively alter how information is distributed between continuous and discrete latent spaces.
\begin{figure}[h]
\floatconts
  {fig:problem_figure}
  {\caption{Latent space of the Joint-VAE (colored by ground-truth sites and learned discrete classes). As continuous capacity increases, the discrete space collapses, merging distinct site clusters and illustrating the sensitivity of the original Joint-VAE to manual capacity tuning, motivating the need for a principled, automatic mechanism to balance continuous and discrete representations.
}}
{\includegraphics[width=0.80\linewidth]{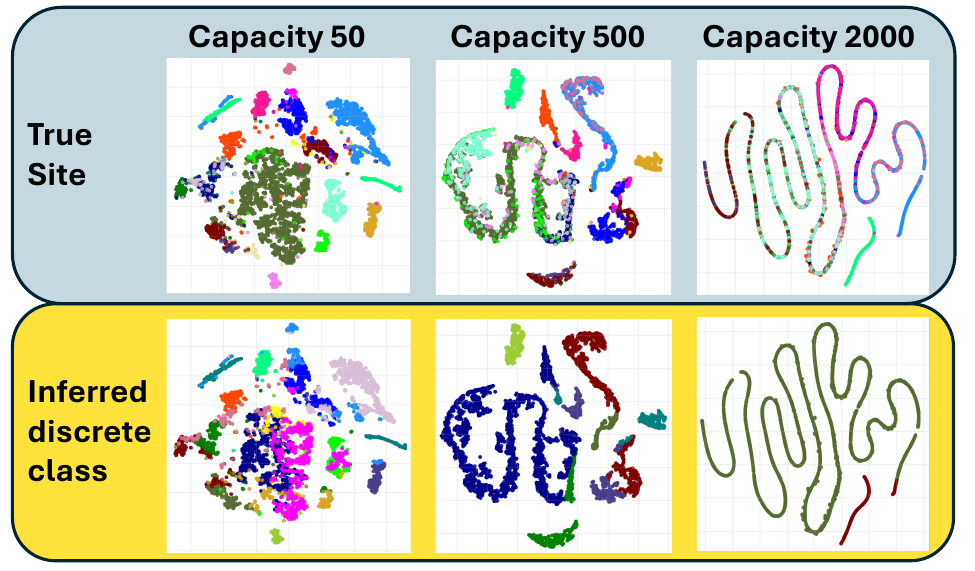}}
\end{figure}
\subsection{Site learning comparison for improved Joint-VAE with annealing}
We then evaluate how well each method recovers the underlying acquisition sites by varying the number of discrete classes from 5 to 25, where 25 corresponds to the true number of sites in the dataset. 
For PCA and the fully continuous VAE baseline, we use the same number of clusters for $k$-means. 
At low numbers of classes, PCA, the continuous VAE, and both Joint-VAE variants perform similarly, likely because the small number of classes forces many true sites to be merged together, effectively averaging across multiple acquisition conditions.
However, as the number of classes approaches the true value, the differences between methods become pronounced. 
Beginning around 20 classes, both PCA and the loss-annealed Joint-VAE show a marked drop in performance, whereas the model-annealed Joint-VAE remains substantially more stable. 
At 25 classes, the ARI achieved by the model-annealed Joint-VAE is significantly higher ($p < 0.05$) than all other methods, indicating that model-level annealing yields far more reliable and robust site. discovery(Fig.~\ref{fig:quant_figure}A).

To better understand what the model learns, we visualize the continuous latent space in two dimensions usint t-SNE (Fig.~\ref{fig:qual_figure}), coloring each point by the discrete assignments inferred by the model. 
The assignments closely match the ground-truth site labels, which are defined by unique combinations of acquisition parameters. 
To further probe this structure, we color the same latent space using the acquisition parameters themselves. 
The latent space separates multi-shell from single-shell acquisitions clearly, and within the single-shell group, the latent space smoothly organizes subjects according to TE, TR, and angular resolution (number of directions). 
These patterns demonstrate that the model discovers a meaningful factorization of acquisition variability: rather than being driven by any single parameter, the structure reflects a nonlinear combination of all acquisition characteristics. 
Importantly, this organization emerges in an unsupervised manner solely through the joint continuous–discrete representation.
\begin{figure}[H]
\floatconts
  {fig:quant_figure}
  {\caption{A) ARI comparison across methods as the number of discrete classes varies. Joint-VAE with model-based annealing and Joint-VAE with loss-based annealing both outperform PCA+$k$-means and VAE+$k$-means as the number of classes approaches the true number of acquisition sites (25), with the model-annealed Joint-VAE showing the most stable and consistently high ARI ($p<0.05$; 1000 bootstrap resamples). B) Sensitivity analysis of the annealing suppression duration. Both ARI and homogeneity remain stable across a wide range, with performance degrading when suppression consumes most of training, indicating robustness to this hyperparameter while highlighting the need for sufficient joint-optimization time.}}
  {\includegraphics[width=0.85\linewidth]{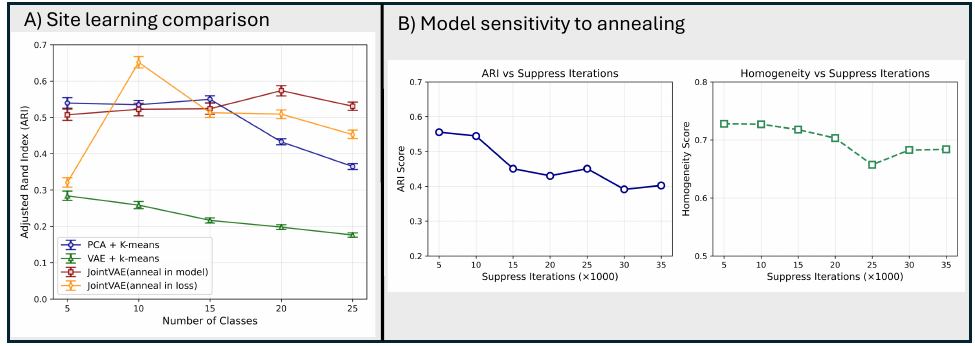}}
\end{figure}
\subsection{Sensitivity of improved Joint-VAE to annealing}
To assess how sensitive the proposed method is to the annealing hyperparameter, we perform a sweep over the number of iterations used for the linear increase of $\lambda$. 
Because $\lambda$ always ranges from 0 to 1, its effect is controlled entirely by the duration of the warm-up schedule. We therefore vary the annealing length from $5,000$ iterations up to the full 35{,}000 training iterations, with the latter effectively suppressing joint optimization for the entire training regime(Fig.~\ref{fig:quant_figure}B). Across most of this range, both ARI and homogeneity remain stable, indicating that the model is relatively robust to the choice of warm-up length. However, when the annealing schedule is extended too far, the model’s ability to recover site structure degrades: prolonged suppression of the continuous channel prevents the network from fully transitioning into the joint learning objective.
\begin{figure}[H]
\floatconts
  {fig:qual_figure}
  {\caption{Left: Data points colored by true acquisition site and model-learned discrete assignments, showing strong correspondence. Right: the same space colored by acquisition parameters using the LAB scientific color space (L = TE, A = TR, B = $\#$directions). Marker size indicates shell value for single-shell data, and symbol type indicates shell combinations for multi-shell data. The latent structure captures multiple acquisition characteristics simultaneously, demonstrating that the model learns diverse sources and groupings of acquisition variability.}}
  {\includegraphics[width=0.9\linewidth]{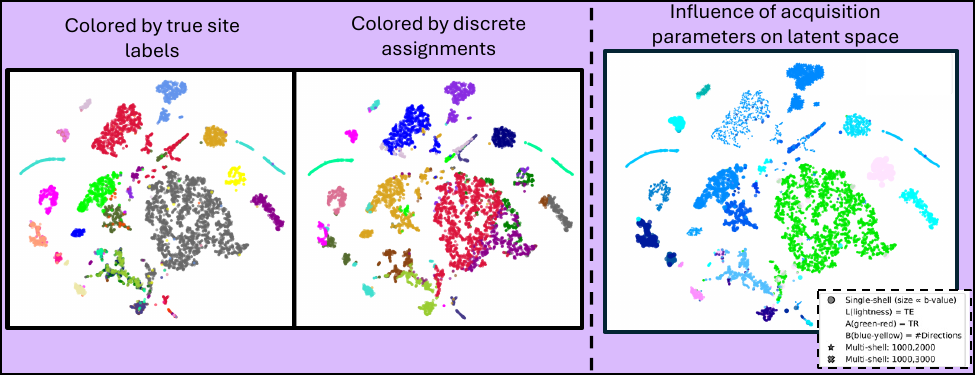}}
\end{figure}

\section{Discussion and Conclusion}

Our findings demonstrate that hybrid continuous–discrete latent spaces provide a useful mechanism for capturing acquisition-related variability in dMRI in a fully unsupervised manner. By explicitly modeling both smooth variation and categorical structure, the Joint-VAE is able to recover meaningful clusters corresponding to scanner and protocol differences. The proposed model-based annealing strategy further offers a principled way to balance the capacities of the continuous and discrete components without manual tuning, enabling stable and reliable site discovery across a wide range of settings. The structure of the learned latent space reinforces that no single acquisition parameter dominates dMRI variability. Instead, the latent representation reveals that TE, TR, angular resolution, b-values, and shell structure all contribute jointly, and the hybrid latent space naturally captures this multi-factor dependence.

A notable limitation is that the latent space is dominated by acquisition-related variation rather than biological differences between subjects. This is expected in a fully unsupervised setting given the strength of acquisition effects in large, heterogeneous datasets, but it highlights an important direction for future work. A semi- or weakly supervised extension—where the discrete latent variable captures acquisition structure in this unsupervised way, while dedicated continuous dimensions are encouraged to encode biological variability—could better separate these factors. Such models would support large-scale, cross-study neuroimaging analyses that account for acquisition heterogeneity while preserving meaningful biological signal.

\clearpage  
\midlacknowledgments{This work was conducted in part using the resources of the Advanced Computing Center for Research and Education at Vanderbilt University, Nashville, TN, and was supported by NIH grant R01EB017230 (PI: Landman) and K01AG073584(PI: Archer). We would also like to thank the Vanderbilt Institute for Clinical and Translational Research (VICTR) for providing computation resources to aid in this research. This work was supported by the Alzheimer’s Disease Sequencing Project Phenotype Harmonization Consortium (ADSP-PHC) that is funded by NIA (U24 AG074855, U01 AG068057 and R01 AG059716). 

This work was also supported by the National Cancer Institute (NCI) grants R01 CA253923 and R01 CA275015
This research was supported in part by the Intramural Research Program of the National Institutes of Health (NIH). The contributions of the NIH author(s) were made as part of their official duties as NIH federal employees, are in compliance with agency policy requirements, and are considered Works of the United States Government. However, the findings and conclusions presented in this paper are those of the author(s) and do not necessarily reflect the views of the NIH or the U.S. Department of Health and Human Services.
The BLSA dataset was supported by the Intramural Research Program of the National Institute on Aging, NIH. Data from the Wisconsin Registry for Alzheimer’s Prevention (WRAP) was supported by NIA grants AG021155, AG0271761, AG037639, and AG054047.

We gratefully acknowledge the efforts of the HABS-HD MPIs: Sid E. O’Bryant, Kristine Yaffe, Arthur Toga, Robert Rissman, and Leigh Johnson, as well as the HABS-HD Investigators: Meredith Braskie, Kevin King, James R. Hall, Melissa Petersen, Raymond Palmer, Robert Barber, Yonggang Shi, Fan Zhang, Rajesh Nandy, Roderick McColl, David Mason, Bradley Christian, Nicole Phillips, Stephanie Large, Joe Lee, Badri Vardarajan, Monica Rivera Mindt, Amrita Cheema, Lisa Barnes, Mark Mapstone, Annie Cohen, Amy Kind, Ozioma Okonkwo, Raul Vintimilla, Zhengyang Zhou, Michael Donohue, Rema Raman, Matthew Borzage, Michelle Mielke, Beau Ances, Ganesh Babulal, Jorge Llibre-Guerra, Carl Hill, and Rocky Vig. Research related to HABS-HD was supported by the National Institute on Aging of the National Institutes of Health under award numbers R01AG054073, R01AG058533, R01AG070862, P41EB015922, and U19AG078109.

Data contributed from the ROS/MAP/MARS studies were supported by grants from the National Institute on Aging: R01AG017917, P30AG10161, P30AG072975, R01AG022018, R01AG056405, UH2NS100599, UH3NS100599, R01AG064233, R01AG15819, and R01AG0-67482, along with support from the Illinois Department of Public Health (Alzheimer’s Disease Research Fund). These data are available at www.radc.rush.edu.
Data were also provided in part by OASIS-4: Clinical Cohort, led by Principal Investigators T. Benzinger, L. Koenig, and P. LaMontagne.

We also acknowledge the National Alzheimer’s Coordinating Center (NACC) database, which is funded by NIA/NIH Grant U24AG072122. Data were contributed by NIA-funded Alzheimer’s Disease Research Centers (ADRCs), including but not limited to: P30AG062429 (PI: James Brewer), P30AG066468 (PI: Oscar Lopez), P30AG062421 (PI: Bradley Hyman), P30AG066509 (PI: Thomas Grabowski), P30AG066514 (PI: Mary Sano), P30AG066530 (PI: Helena Chui), P30AG066507 (PI: Marilyn Albert), P30AG066444 (PI: John Morris), and numerous others listed in full above. Data collection and sharing for this project was funded by the Alzheimer's Disease Neuroimaging Initiative (ADNI) (National Institutes of Health Grant U01 AG024904) and DOD ADNI (Department of Defense award number W81XWH-12-2-0012). ADNI is funded by the National Institute on Aging, the National Institute of Biomedical Imaging and Bioengineering, and through generous contributions from the following: AbbVie, Alzheimer’s Association; Alzheimer’s Drug Discovery Foundation; Araclon Biotech; BioClinica, Inc.; Biogen; Bristol-Myers Squibb Company; CereSpir, Inc.; Cogstate; Eisai Inc.; Elan Pharmaceuticals, Inc.; Eli Lilly and Company; EuroImmun; F. Hoffmann-La Roche Ltd and its affiliated company Genentech, Inc.; Fujirebio; GE Healthcare; IXICO Ltd.; Janssen Alzheimer Immunotherapy Research \& Development, LLC.; Johnson \& Johnson Pharmaceutical Research \& Development LLC.; Lumosity; Lundbeck; Merck \& Co., Inc.; Meso Scale Diagnostics, LLC.; NeuroRx Research; Neurotrack Technologies; Novartis Pharmaceuticals Corporation; Pfizer Inc.; Piramal Imaging; Servier; Takeda Pharmaceutical Company; and Transition Therapeutics. The Canadian Institutes of Health Research is providing funds to support ADNI clinical sites in Canada. Private sector contributions are facilitated by the Foundation for the National Institutes of Health (www.fnih.org). The grantee organization is the Northern California Institute for Research and Education, and the study is coordinated by the Alzheimer’s Therapeutic Research Institute at the University of Southern California. ADNI data are disseminated by the Laboratory for Neuro Imaging at the University of Southern California.

During the preparation of this work, the authors used GitHub Copilot and ChatGPT to assist with code generation, debugging, editing, and sentence restructuring. All AI-generated content was reviewed and edited by the authors, who take full responsibility for the final content of the publication.
}

\bibliography{midl-samplebibliography}
\end{document}